# Design and Development of a Prototype Robotic Gripper

\* Ramish and Farah Kanwal
*Department of Mechanical Engineering*
*NED University of Engineering & Technology*
*Karachi, Pakistan*
rameshchouhan66@yahoo.com
farah_kanwal@live.com

Siraj Ali and Nida Ali
*Department of Mechanical Engineering*
*NED University of Engineering & Technology*
*Karachi, Pakistan*
engr_siraj@hotmail.com
engineer.nida@hotmail.com

*Abstract* - **Robotic grippers are widely used in industries for handling objects. This paper presents the procedure of design and fabrication of a stepper motor controlled robotic gripper to be used in industries for handling small objects.**
**The gripper has one degree of freedom for linear motion and one degree of freedom for rotational motion. A linear actuator has been employed to give translational motion to the gripper. Based on the design requirement data, a kinematic analysis has been done to accurately predict the design parameters useful in proper selection of motor and linear actuator.**

## I. INTRODUCTION

In this paper, design and development of a planar gripping device is presented which is capable of picking and lifting certain regular shaped objects like cubic and cylindrical boxes in production line. Robotic grippers are widely used in industries for handling objects. One of the most common grasping methods used in robotic grippers is Mechanical gripper, where friction or the physical configuration of the gripper retains the object. [1]

The style of robot jaws that is used, play major roll to determine the force requirement in gripper application. There are two types of jaws: friction or encompassing. Friction grip jaws rely totally on the force of the gripper to hold the part meanwhile encompassing jaws add stability and power by cradling the part. In other hand, friction grip pick up the desired item with "hand" completely flat like paddles meanwhile encompassing grip use "fingers" spreads and wrapped around the items [2].

The economic aspect is one of the most important to be considered during the whole development. The simple design of this gripper ensures the ease of manufacturability and reliability of operation.

## II. GRIPPER DESIGN

### A. Mechanical Design

For designing the overall structural layout of this gripper, several aspects had been considered such as simplicity of design, simple and low cost manufacturing, reliability, durability & good aesthetics.

Our design is based on following guidelines [3]:

1) Gripper weight should be minimized. This favors the robot to accelerate more quickly

2) Grasping of objects should be secure: This allows the robot to run at higher speeds in zig-zag profile thereby reducing the cycle time.

3) Completely encompass the object with the gripper: This is to help hold the component securely.

4) Grasp the object without deformation: The objects are easily deformed and so care should be taken during grasping these objects.

5) Minimize finger length: The longer the fingers of the gripper the more they are going to deflect during grasping an object.

The schematic of the gripper is shown in Fig. 1. The curved fingers have been designed to give an effective grasping to certain prismatic objects such as cylinders.

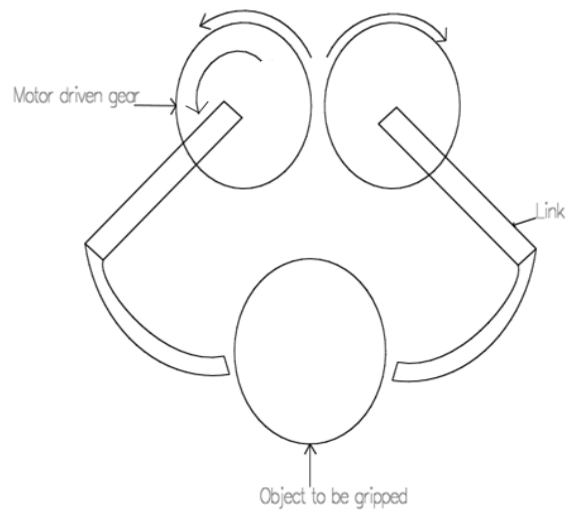

Fig. 1 Schematic of Robotic Gripper

Each link is attached with the spur gear. A DC motor is coupled directly with one of this gear with step down configuration. Relatively simpler assembly ensures smooth & reliable operation. Safe and reliable operation is ensured by the use of minimum number of moving parts. The design is robust because only by changing the end effectors, it can be made to grip any shape of object. The performance of this device largely depends on the design of these end effectors.

Fig. 2 shows the assembled view of the gripper. Curved fingers 1 are bolted to the slotted T shaped beam 4. This beam is rigidly attached to the centre of gears 3 and 7 through flat rectangular links. This assembly is sandwiched between two flat rectangular plates 2. The DC motor 6 is coupled to gear 7 through a pinion 5 and mounted on the plate. The curved fingers and other parts were machined from aluminium. Aluminium was selected for this prototype gripper because of its good machinability properties. However, for handling heavy objects in industrial environments, steel or other such high strength material can be used.

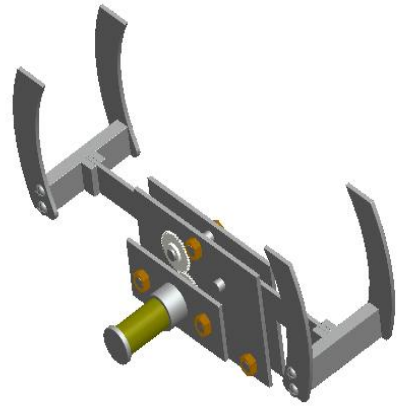

Fig. 3 Solid Model of Gripper

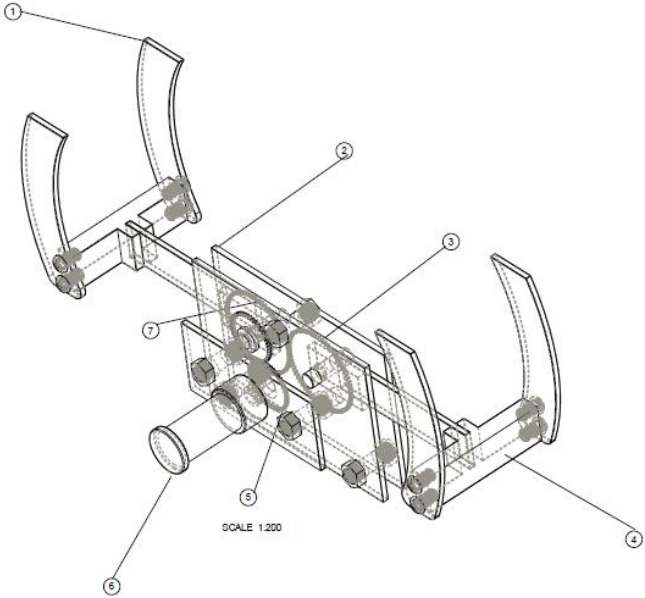

Fig. 2 Wire frame model of gripper

The end effectors are not shown here, but it can be installed at the tip of the curved fingers as per requirement. The flat and slender members used in the structure of this gripper ensure low inertia and less power consumption in the idle movement of the arm. However, the stiffness of these slender links must be evaluated assuming them as cantilever beams. [4]

The whole assembly can be installed on the linear actuator or other such driving mechanisms to give a to and fro movement to gripper. This will be helpful where it has to be mounted to grip and lift certain objects. Such configuration is very useful in pick and place jobs like picking of packaging from the industrial conveyors and placing them at the other place.

An assembled solid model of the gripper is shown in the Fig. 3.

### B. Selection of Motor for Drive

Selection of proper drive for the gripper is very crucial for reliable operation. Different factors that might affect the choice of drive would be the accuracy, availability of power supply, cost etc. Servo drives are recommended for applications where accuracy of positioning is required. Stepper motors are next available option for such intermittent operation where motor has to turn a certain angle in one direction, hold the object with certain torque (stalling condition) and turn again in the opposite direction. Another option is to use a simple DC motor with some modifications, incorporated in our design and discussed in the next paragraph.

In this design, a 12V DC motor drives the gear 7 which in turn drives the assembly to open the jaw and grip the object. The gripping force depends upon the stalling torque of the DC motor being used. Stalling the DC motor for a long time can cause over heating of the windings of motor. To avoid this problem, a PWM (Pulse Width Modulating) circuit was made to operate the motor. PWM as it applies to motor control is a method of exciting the circuit through a succession of pulses rather than a continuous analog signal as shown in Fig. 4. By varying the pulse width, the controller regulates energy flow to the motor.

### III. GRIPPING FORCE AND MOTOR TORQUE CALCULATION

The analysis of the gripping force and motor torque are given below.

Tolouei-Rad and Kalivitis [5] used a simple planar model to calculate the gripping force. A similar model has been used to calculate the torque of the motor required to turn the gear, a simplified mathematical model is built with all the attached parts assumed as point masses. Then, elementary equations of Dynamics have been applied to model the dynamic behaviour of this gripping device.

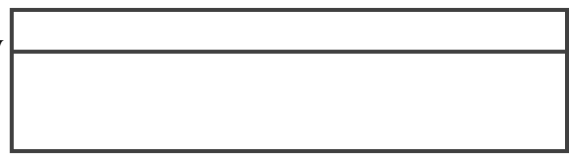

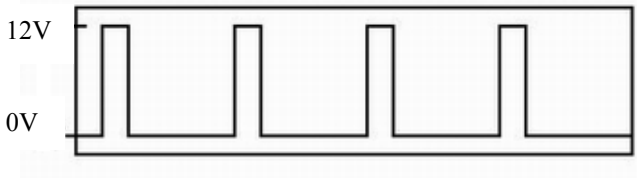

Fig. 4 PWM of a 12V DC Signal

For making the analysis simple, following assumptions have been taken [6],

1. Rigid-body models with point contacts between the fingertips and the grasped object, often in 2-D
2. Linearized (instantaneous) kinematics
3. Quasistatic analysis (no inertial or viscous terms)
4. No sliding or rolling of the fingertips
5. No cases with redundant degrees of freedom and no over-constrained grasps.
6. Assumed full knowledge of object and contact states, with no consideration for using sensory information during manipulation

Consider the free body diagram shown in Fig. 5. The gripper holds the object of mass "m". The torque required by the motor at point O can be found by applying conditions of equilibrium.

T = minimum holding torque or no speed torque that the motor must apply to grasp the object (N-m)
$\mu$ = coefficient of friction between gripper finger & object surfaces.
R = Normal Reaction between gripping fingers and the object to be grasped (N)
g = acceleration due to gravity (m/s$^2$)
a and b = moment arms for the forces as shown in Fig. 5

Summing forces in the y-direction on object,
$\uparrow + \Sigma F_y = 0;$
$2\mu R = mg$
$R = mg/2\mu$

Summing moments about point O,
$+\Sigma M_O = 0;$
$-Rb + \mu Ra - T' + T = 0$
$T = T' + Rb - \mu Ra$ ......................................... (1)

For T' summing moments about point O',
$+\Sigma M_{O'} = 0;$
$Rb - \mu Ra - T' = 0$

$T' = Rb - \mu Ra$ ......................................... (2)

Putting the value of T' from (1) into (2)
$T = 2(Rb - \mu Ra)$

The parameters a and b would depend upon the size of the object to be grasped.

IV. APPLICATIONS

Robot models are created with specific applications or processes. Different applications have different requirements. They make work easier and save time and money.

They are used in industries as welding robots, material handling robots, painting robots, palletizing robots and assembly robots. They have a wide application in medical surgeries. The normal movements of a surgeon is converted by a remote manipulator into the movements of robotic gripper to perform the actual surgery on the patient. Also they are used on space shuttle orbiters for inspection purposes.

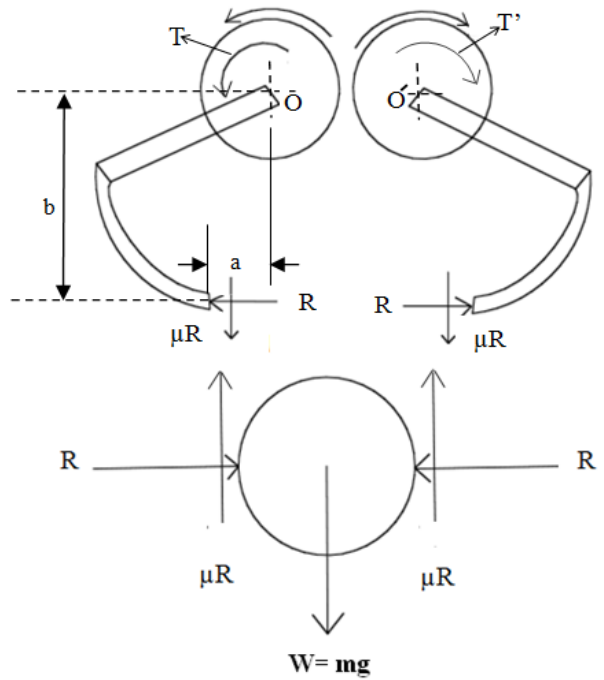

Fig. 5 Free body diagram of gripping assembly

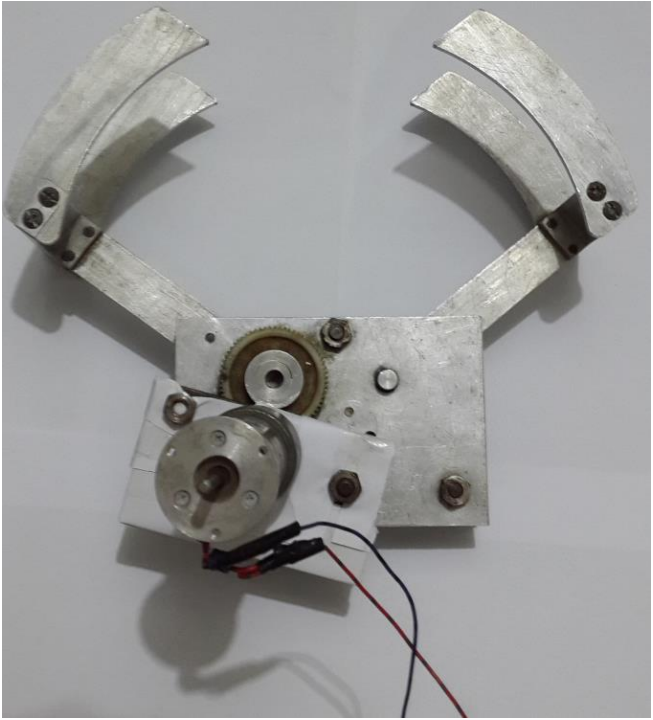

Fig. 6 Gripping assembly

## V. Conclusion

The robotic gripper has been designed and fabricated taking into account the economics, simplicity and reliability. Suitable end effectors can be installed for particular application. They should be designed according to the shape of the object to be gripped. We have installed the motor in our prototype which was easily available. However, for large scale application the given numerical method could be used for selecting the right motor.

### Acknowledgment

We express our deepest gratitude to Prof. Dr. Mubashir Ali Siddique, Chairman of Mechanical Engineering Department and Prof. Dr. Amir Iqbal, Chairman of Industrial and Manufacturing Engineering Department who allowed us to use the Workshop facility. We would like to thank the workshop staff for their guidance and support.